\documentclass[10pt,twocolumn,letterpaper]{article}

\usepackage{float}

\usepackage{cvpr}
\usepackage{times}
\usepackage{epsfig}
\usepackage{graphicx}
\usepackage{amsmath}
\usepackage{amssymb}
\usepackage{subcaption}
\usepackage{abbrev_ds}
\newcommand{\specialcell}[2][c]{%
\begin{tabular}[#1]{@{}c@{}}#2\end{tabular}}
\usepackage{bm}
\usepackage[percent]{overpic}
\usepackage{minibox}
\usepackage{caption}
\usepackage{color}

\newcommand{\Fig}{Fig.}

\usepackage[pagebackref=true,breaklinks=true,letterpaper=true,colorlinks,bookmarks=false]{hyperref}

\cvprfinalcopy 

\begin{document}

\title{Optical Flow with Semantic Segmentation and Localized Layers}

\author{Laura Sevilla-Lara$^1$ \quad Deqing Sun$^{2, 3}$ \quad Varun Jampani$^{1}$ \quad Michael J. Black$^1$\\ 
  \begin{tabular}{c c}
& \\
    $^1$MPI for Intelligent Systems & $^2$NVIDIA, $^3$Harvard University \\
    {\tt\small \{laura.sevilla, varun.jampani, black\}@tuebingen.mpg.de} & {\tt\small deqings@nvidia.com}
  \end{tabular}     
       }

\twocolumn[{%
\renewcommand\twocolumn[1][]{#1}%
\maketitle
\begin{center}
    \newcommand{\teaserwidth}{1.\textwidth}
     \newcommand{\Shiftleft}{\hspace{-1em}}
\vspace{0.in}
    \centerline{
    \begin{tabular}{c}%
\includegraphics[width=17.45cm]{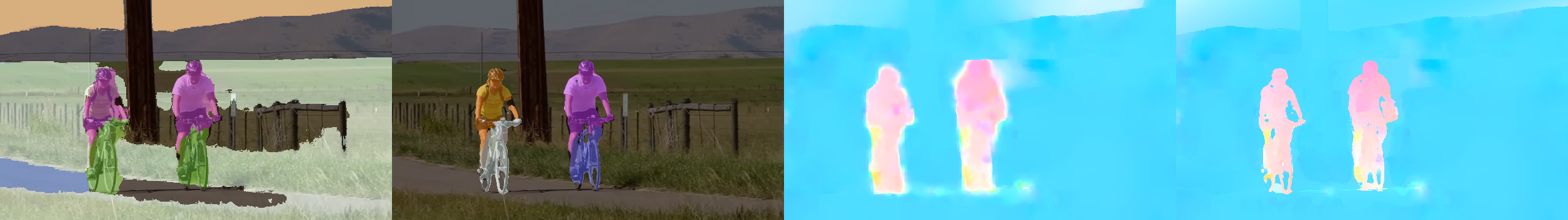} \\%
 \hspace{-1.1cm}  {\small (a) Initial segmentation~\cite{ChenPKMY14}} \hspace{1.4cm}  {\small (b) Our segmentation }   \hspace{1.4cm} {\small (c) DiscreteFlow~\cite{Menze2015GCPR}}  \hspace{1.4cm} {\small (d) Semantic Optical Flow}
 \end{tabular}}
\vspace{-0.07in}
    \captionof{figure}{
     (a) Semantic segmentation breaks the image into regions such as road, bike, person, sky, etc.
(c)  Existing optical flow algorithms do not have access to either the
 segmentations or the semantics of the classes.
(d) Our semantic optical flow algorithm computes motion
differently in different regions, depending on the semantic class
label, resulting in more precise flow, particularly at object boundaries.
(b) The flow also helps refine the segmentation of the
foreground objects.  }
        \label{fig:teaser}
\end{center}
}]

\maketitle

\begin{abstract}


Existing optical flow methods 
make generic, spatially homogeneous, assumptions about the spatial structure of the flow. 
In reality, optical flow varies across an image depending on object class.
Simply put, different objects move differently.
Here we exploit recent advances in static semantic scene segmentation to segment the image into objects of different types.
We define different models of image motion in these regions depending on the type of object.
For example, we model the motion on roads with homographies, vegetation with spatially smooth flow, and independently moving objects like cars and planes with affine motion plus deviations.
We then pose the flow estimation problem using a novel formulation of {\em localized layers}, which addresses limitations of traditional layered models for dealing with complex scene motion.
Our {\em semantic flow} method achieves the lowest error of any published monocular method in the KITTI-2015 flow benchmark and produces qualitatively better flow and segmentation than recent top methods on a wide range of natural videos.



\end{abstract}

\section{Introduction}
\label{sec:intro}

The accuracy of optical flow methods is improving steadily, as evidenced by results on several recent datasets \cite{Butler:2012:Sintel,Geiger2013IJRR}.
However, even state-of-the-art optical flow methods still perform poorly with fast motions, in areas of low texture, and around object (occlusion) boundaries (\Fig~\ref{fig:teaser} (c)).
Here we address these issues and improve the estimation of optical flow by using semantic image segmentation.
Like flow, the field of semantic segmentation is also making rapid progress, driven by convolutional neural networks (CNNs) and large amounts of labeled  data.
Here we use a state-of-the-art method \cite{ChenPKMY14}  (\Fig~\ref{fig:teaser} (a)) and find that existing semantic segmentation methods, while not perfect, are good enough to significantly improve flow estimation.

We use semantic image segmentation in multiple ways.
First, it provides information about object boundaries.
Second, different objects move differently; roads are flat, cars move independently, and trees sway in the wind.
This means that our prior expectations about the image motion should vary between regions with different class labels.
Third, the spatial relationships between objects provide information about the relative local depth ordering of regions.
Reasoning about depth order is typically challenging and we use the semantics to
simplify this, improving flow estimates at occlusion boundaries.
Fourth, object identities are constant over time, providing a cue that we exploit to encourage temporal consistency of the optical flow.

\begin{figure}[t]
\centerline{\includegraphics[width=3.25in]{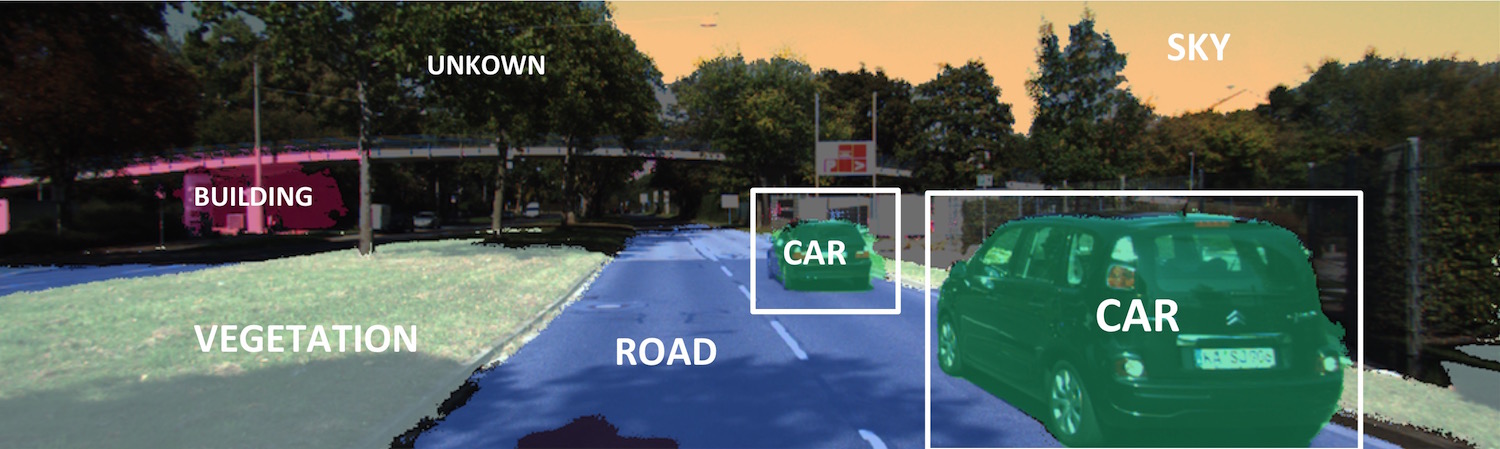}}
\caption{{\bf Localized layered model.} An image is segmented into semantic regions (color coded).  Different regions are assigned different motion models.
Independently moving objects are shown with a box around them.  These regions require reasoning about occlusion because such objects move in front of the background.
Within each such region, we make the assumption that two motions are present (the background and the foreground object).
The formulation is similar to previous layered models but here the spatial extent of each layer may vary.
}
\label{fig:layers}
\end{figure}

To model complex scene motions and to deal well with motion boundaries, we adopt a layered approach
\cite{Ayer:1995:LRMDL,Darrell:1991:REMLM,Hsu:1994:ICPR,Jepson:1993:MMOF,Jojic:2001:LFS,Sun:CVPR:2013,Wang:1994:RMIL,Weiss:1997:SL,Wulff_CVPR_2015}.
Layered models, however are typically global and cannot represent complex occlusion relationships.
There have been attempts to formulate locally layered models \cite{Black:CVPR:1996,Sun:CVPR:2014}, but these methods are still spatially homogenous.
Here we propose a new model of {\em localized layers} in which the number of layers in the scene varies spatially.
Any pixel of the scene may belong to one or more layers and these layers may have varying spatial extent.
Local layered models are used as needed to capture the motion of relevant objects.
In regions corresponding to objects that can move, we may find two motions -- the foreground motion of the object against a background motion.
Here we use local two-layer models.
Rather than a small number of global layers, the result is a patchwork of smaller layered regions on top of background regions as illustrated in \Fig~\ref{fig:layers}.
The approach keeps the complexity and optimization manageable by using at most two layers within any patch.
And because we can use as many patches as needed, the approach can model complex motions.
This adaptive, spatially heterogeneous approach extends layered models to more complex scenes and uses them where they are most valuable.

Each layer or region is represented by a motion model and the type of model varies depending on the semantic label of the region.
For regions that are likely to be planar we model their motion with a homography; this includes roads, sky, and water.
For regions corresponding to independently moving objects, we treat their motion as affine but allow it to deviate from this assumption; these classes include objects like cars, planes, boats, horses, bicycles, and people.
There are still other classes like vegetation and buildings that are diverse in their 3D shape and motion and are consequently not well modeled by a simple parametric motion.
Consequently, we model these classes with a classical spatially varying dense flow field.
The motion of the scene is then described by composing the motions of all the semantic regions (\Fig~\ref{fig:composing}).

We call the algorithm \emph{semantic optical flow (SOF)} because it exploits scene semantics to improve flow estimation.
The approach achieves the lowest error on the KITTI-2015 flow dataset \cite{Menze2015CVPR},  when compared with all published monocular flow methods.\footnote{The most accurate methods use stereo motion sequences and exploit the stereo to estimate scene structure.}
We also test the method on a challenging range of sequences from the Internet.
There are several reasons for the improvements.
First our motion models provide a form of long-range regularization in areas like roads.  Since these are well modeled by a homography, accuracy improves.
Second, this region-based regularization helps flow estimation in homogeneous regions, which contain few motion cues.
Third, the localized layer formulation improves the segmentation and flow around motion boundaries.
Key here is that the object segmentation gives a good initialization for layered flow segmentation and 
gives a good hypothesis for which surface is in front and which is behind; this improves occlusion estimation.

While we focus on improving optical flow, we note that motion can also help with scene segmentation.
While current semantic segmentation methods are good, they still struggle to separate object boundaries from appearance boundaries  (\Fig~\ref{fig:teaser} (a)).
Layered optical flow estimation segments the region and provides additional information about object boundaries (\Fig~\ref{fig:teaser} (b)).
When computed over several frames, this segmentation can be quite precise.

In summary, we make two contributions. First, we present the first optical flow method that uses semantic information about scenes, objects, and their segmentation, producing the lowest error among all monocular methods on the KITTI flow benchmark. Second, we show how layered optical flow estimation can be extended to cope with complex scenes.
Our results confirm that knowing what and where things are helps the estimation of how they move.

\section{Related Work}
\label{sec:related_work}

{\bf Motion estimation and segmentation.} 
There is a long history of simultaneously estimating optical flow and its segmentation \cite{Memin:TIP:1998,murray-buxton}.
Many methods focus on segmentation using motion information alone; we do not consider these here.
More relevant are methods that use image segmentation to aid optical flow.
Previous work~\cite{Black:IEEE:1996, Yang_2015_CVPR} segments the scene into patches according to color or other cues, and then fits parametric flow models within these.
Like us they vary the type of model in each region but we go beyond this to use semantic information to determine the appropriate model.
Sun et al.~\cite{Sun:CVPR:2014} first segment the scene into superpixels and then reason about the occlusion relationships between neighboring superpixels (cf. \cite{Yamaguchi:CVPR:2013}).
These methods are generic in the sense that they do not know anything about the objects being segmented but rather seek a partitioning of the scene into coherently moving regions.

{\bf Combining flow models.}
Here we use different flow models to represent the motion of different parts of the scene.
These are combined within our localized layer formulation to define the flow for the whole image.
Previous work has explored the combination of different flow algorithms  \cite{FusionFlow,Brostow}.
Irani and Anandan~\cite{IraniA98} develop a theory for modeling motion in general scenes with varying levels of complexity.
The above methods, however, are generic in the sense that they do not use any semantic information about objects to select among the possible models.

{\bf Occlusion reasoning and figure-ground.}
One goal of optical flow estimation is the detection of motion discontinuities that may signal the presence of an object (surface) boundary (see \cite{Thompson:1998:EDO} for an overview).
Previous methods focus on generic constraints without taking into account object-specific information \cite{Black:2000:DTMB,Stein:2009:OBM,Sundberg:2011:OBD,Thompson:1998:EDO}.
In these cases the goal is to detect boundaries that may be useful later for object detection.
We turn this around by performing object detection and then using this to detect motion boundaries more accurately.


{\bf Layered optical flow.}
Layered flow estimation has a long history
\cite{Ayer:1995:LRMDL,Darrell:1991:REMLM,Hsu:1994:ICPR,Jepson:1993:MMOF,Jojic:2001:LFS,Wang:1994:RMIL,Weiss:1997:SL}
and recent improvements have made the approach more competitive on standard benchmarks \cite{Sun:CVPR:2013} and more computationally tractable \cite{Wulff_CVPR_2015}.
The most recent work integrates image segmentation cues with motion cues to produce an accurate segmentation at motion boundaries.
In particular, we build on \cite{Sun:CVPR:2013}, which uses a fully connected graphical model (cf. \cite{Krahenbuhl2011FCRF}) to exploit long-range image cues for layer segmentation.
Unlike previous work, we apply the model locally within image patches around segmented objects that can move.




Traditional layered models have limitations and are most applicable to simple scenes with a small number of moving objects.
Occlusion relationships in the world are complex and 2D motion layers are too restrictive to capture the 3D spatial occlusion relationships in real scenes.
Also, while the depth order of layers is important, this may be ambiguous in two frames~\cite {Sun:2012:LSOT}.
Reasoning about layer depth order is combinatorial ($K!$ for $K$ layers), which becomes infeasible in realistic scenarios.
To address these issues, locally layered models of motion have been proposed \cite{Black:CVPR:1996,Sun:CVPR:2014}.
These models, again, are generic and do not know about objects.
Here we find the problem of depth order reasoning is often simplified when we have semantic information.
For example, we assume that independently moving objects like cars are in front of static objects like roads.
When the assumption holds, as it often does, this simplifies layered flow estimation and produces accurate motion boundaries.

Several methods decompose scenes into layers corresponding to objects \cite{Jepson:2002:LM,Jojic:2001:LFS,Kumar:2008:LLMSV,Wang:2009:SOMT,ZhouT03}.
What these methods mean by ``object,'' however, is a region of the image that moves coherently and differently from the background; there is no notion of what this object is.
In contrast, Isola and Liu~\cite{Isola2013CVPR} represent static images of scenes as a patchwork of objects layered on top of each other but
they do not consider image motion.

{\bf Video segmentation.} There is significant and increasing interest in the field \cite{Galasso:ICCV:13,Grundmann:2010:HGS,Liu2015CVPR,Ochs:PAMI:14,Ren:CVPR:13,Corso:ECCV:12} but
the definition of the problem varies between identifying coherent motions or coherent objects regions.
Like the approaches above, these methods are generic in that they focus on bottom-up analysis of regions and motion.
They typically use optical flow as a cue to track superpixels over time to establish temporal coherence.
They usually do not use high-level object recognizers or try to improve optical flow.
Taylor \etal~\cite{Taylor:2013:SemanticVideoSegmentation} incorporate object detections and use temporal information to reason about occlusions to improve their segmentation results, but do not compute optical flow.
 Lalos \etal \cite{Lalos:2010:ObjectFlow} compute optical flow for an object of interest using a tracking-by-detection approach.
Unlike us, they only estimate object displacement (not full flow), ignore background motion, and do not take object identity into account.

{\bf Semantic segmentation in other low-level vision problems.} Object class influences the way things move, but also influences their shape. Recent work uses semantic segmentation
to resolve ambiguities in stereo~\cite{Guney2015CVPR}, to guide 3D reconstruction~\cite{hane2013joint,Ladický2011}, and to constrain the motion of the 3D scene by enforcing class label coherence over time~\cite{scharwachter2014stixmantics}.

\section{Model and Methods}
\label{sec:solution}

Using a semantic segmentation of the scene allows us to model the motion of different regions of the image differently.
We define the motion in the scene compositionally in terms of the motion of the regions.
Below we discuss how we compute the motion for each segmented region and then how we combine these into a coherent flow field.

{\bf Classes.}
We define three classes of objects (Things, Planes, and Stuff) that exhibit different types of motion (see \Fig~\ref{fig:layers}).
{\em (1) Things}~\cite{Adelson:1991,heitz2008learning} correspond to objects with a defined spatial extent, that can move independently, are typically seen in the foreground and may be rigid or non-rigid.
Things include aeroplane, bicycle, bird, boat, bus, car, cat, cow, dog, horse, motorbike, sheep, train and person. 
{\em (2) Planes} are regions like `roads' that have a broad spatial extent, are roughly planar, and are typically in the background.
Other classes that we treat as planes are `sky' and `water'.  Water is treated as a plane because the air/water boundary is often planar.
{\em (3) Stuff}~\cite{adelson2001seeing} corresponds to classes that exhibit textural motion or objects like `buildings' and `vegetation' that may have a complicated 3D shape, exhibit complex parallax, and for which we have no compact motion representation.
Regions of unknown class are modeled as Stuff.

\subsection{Preprocessing}

{\bf Segmentation.}
%
%
We used Caffe~\cite{Jia:2014} to train the semantic segmentation model DeepLab~\cite{ChenPKMY14}, substituting all fully-connected layers in the VGG network~\cite{Simonyan14c} with convolutional layers. We modified the output layer to predict the 22 classes described above and used the atrous~\cite{mallat2008wavelet} algorithm to get denser predictions. We initialized the network with the VGG model and fine-tuned it with standard stochastic gradient descent using a fixed momentum of 0.9 and weight decay of 0.0005 during 200K iterations. The learning rate is 0.0001 for the first 100K iterations and is reduced by 0.1 after every 50K steps. To improve performance~\cite{ChenPKMY14,Krahenbuhl:2012:ENL} we used a densely connected conditional random field (Dense-CRF). The unaries are the CNN output and the pairwise potentials are a position kernel and a bilateral kernel with both position and RGB values. The standard deviation of the filter kernels and their relative weights are cross-validated. The inference in the Dense-CRF model is performed using 10 steps of meanfield. To train the network, we selected 22 of the 540 classes from the Pascal-Context dataset~\cite{mottaghi_cvpr14}.

\begin{figure}[t]
\centerline{\includegraphics[width=\columnwidth]{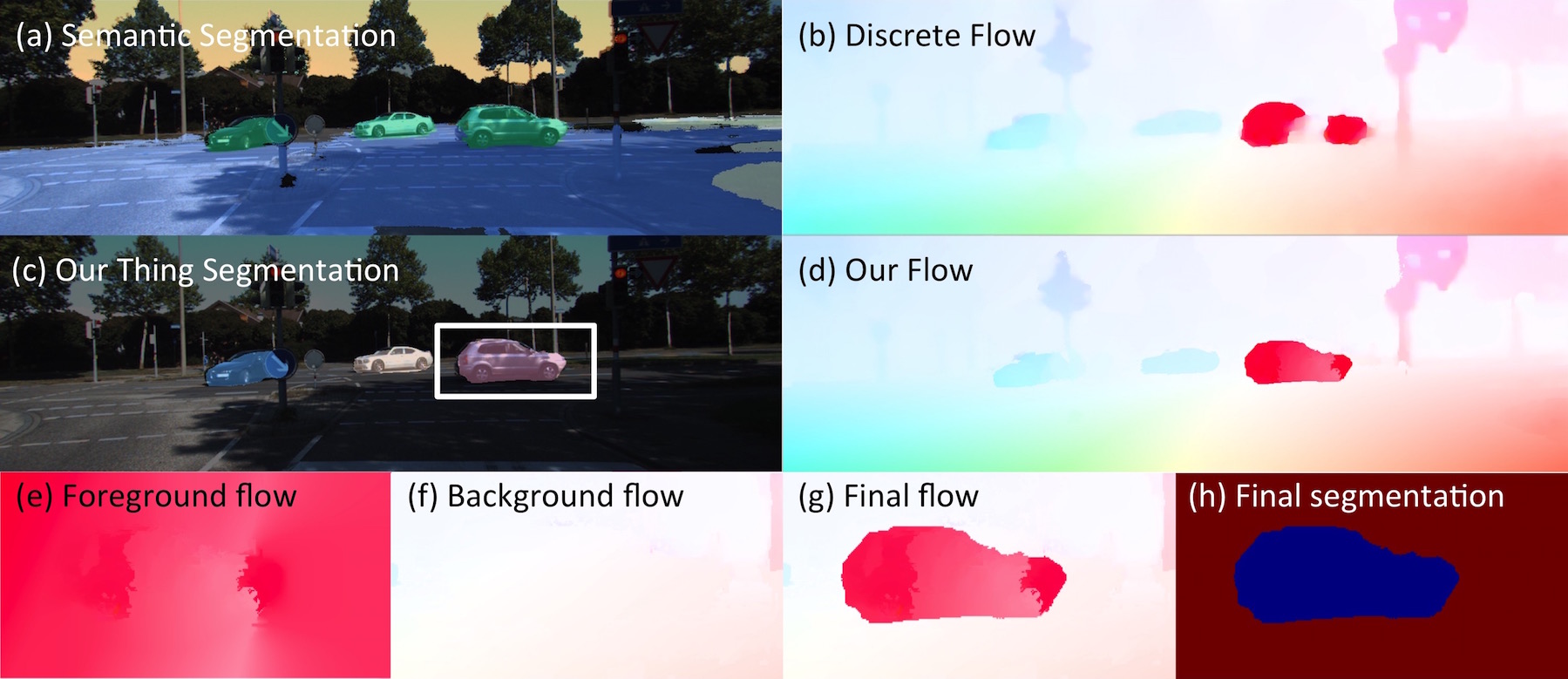}}
\caption{{\bf The method in pictures.}
(a) Image with the segmentation into road (blue), car (green), sky (yellow), grass (grey), and ``unknown'' (clear) superimposed.
(b) Initial dense flow computed with DiscreteFlow~\cite{Menze2015GCPR}.
The following images show intermediate results in the extracted car region. 
(c) Our final Thing segmentation.  (d) Our final flow.
(e) Estimated foreground motion. (f) Estimated background motion.  (g) Estimated flow for the localized region.  (h) Final layer segmentation (blue is foreground).
}
\label{fig:method}
\end{figure}

{\bf Thing matching.}
Given the segmentation in each frame, we compute connected components to obtain regions containing putative objects (Things). 
Regions smaller than 200 pixels are treated as Stuff.
For each Thing found in the first frame, we find its corresponding region in subsequent frames and create a bounding box for layered flow estimation that fully surrounds the object regions across all frames.
This defines the spatial extent of the layered flow estimation  (\Fig~\ref{fig:method}).
Below we estimate the flow of Things using $T=5$ frames at a time unless otherwise stated. 
Figure~\ref{fig:layers} shows a few Thing regions in one frame. If a Thing region is not found over the entire sub-sequence, it is treated as Stuff.

{\bf Initial flow.}
We also compute an initial dense flow field, $\hat{\mv{u}}$ using the DiscreteFlow method~\cite{Menze2015GCPR} based on~\cite{EpicFlow}. 
We use this in several ways as described below.

\subsection{Motion Models}

{\bf The motion of Planes.}
We model planar regions using homographies.
Given the initial flow vectors $\hat{\mv{u}}(\mv{x})$, $\mv{x}\in{R_i}$ in region $i$, we use RANSAC to robustly estimate the parameters, $\mv{h}_i$ of the homography.
The planar motion then defines the flow $\mv{u}_{\textrm{Plane}}(\mv{x}; \mv{h}_i)$ for every pixel $\mv{x}\in R_i$.

{\bf The motion of Stuff.}
For Stuff we have no class-specific motion model and set the flow in every Stuff region $i$ to be the initial flow; that is $\mv{u}_{\textrm{Stuff}}(\mv{x}) = \hat{\mv{u}}(\mv{x})$ for $\mv{x}\in R_i$.

{\bf The motion of Things.}
In Thing regions we expect occlusions and disocclusions, complex geometry, and deformations.
Thus, we assume the motion of a Thing can be described as affine plus a smooth deformation from affine.
This may sound restrictive but we build on the work of \cite{Sun:CVPR:2013}, where they show positive results applying this motion model to the entire scene.
Our Thing regions are much smaller than the entire scene, and the motion within this region is more likely to satisfy the assumptions.
We allow the motion of Things to deviate from affine and the amount of deviation depends on the object class.
For example, cars are more rigid than people and their motion is more affine.
Consequently we assume that the motion of cars will be more affine and penalize deviations from this assumption more. 

While we are interested in the motion of the Thing, because we assume Things are in front of backgrounds, it is actually important to also consider the motion of the background.
Specifically, estimating an accurate foreground segmentation requires that we reason about the motion of both foreground and background.
We do this using a local layered model based on \cite{Sun:CVPR:2013}. 

\begin{figure*}[th]
\begin{align}
\label{eq:fg}
E_{\textrm{Thing}}(\mv{u},\! \mv{v}, \! \mv{g}, \Theta; \! \mv{I}, \hat{\mv{g}}) \!=\!& \sum^{2}_{k=1} \Big \{ \sum^{T-1}_{t=1} \{ E_{\textrm{data}}(\mv{u}_{tk}, \! \mv{v}_{tk}, \! \mv{g}_{tk}; \!  \mv{I}_{t}, \mv{I}_{t\!+\!1})
 \!+\! \lambda_{\textrm{motion}} E_{\textrm{motion}}(\mv{u}_{tk},\! \mv{v}_{tk}, \! \mv{g}_{tk}, \Theta_{tk}) \\
 &\!+\! \lambda_{\textrm{time}} E_{\textrm{time}}(\mv{u}_{tk},\! \mv{v}_{tk}, \! \mv{g}_{t, k}, \mv{g}_{t\!+\!1, k}) \}
 \!+\! \sum^{T}_{t=1} \{ \! \lambda_{\textrm{layer}} E_{\textrm{layer}}(\mv{g}_{tk}; \hat{\mv{g}}_{tk})\!+\! \lambda_{\textrm{space}} E_{\textrm{space}}( \mv{g}_{tk} ) \} \Big \}.
 \vspace{-5in}
\nonumber
\end{align}
\end{figure*}

Formally, given a sequence of images $\{ {\bf I}_t, 1\leq t \leq T \}$, we want to jointly estimate the motion $(\mv{u}_{tk}, \mv{v}_{tk})$ for every pixel, in each layer, at every frame, as well as group pixels that move together into layers denoted by $\mv{g}_{tk}$, where $k\in\{1,2\}$.
We only consider two layers, and thus we only need to estimate the foreground segmentation, $\mv{g}_{t1}$, as the background layer is constant. We formulate the local layered energy term (Eq.~\ref{eq:fg}) similar to Sun \etal with some modifications described below and refer the reader to \cite{Sun:CVPR:2013} for further details.
The method estimates the motion of both layers and the segmentation of the foreground region.

The general formulation incorporates occlusion reasoning in the motion estimation using layered segmentation ({data term}), enforces temporal consistency of layer segmentation ({time term}) according to the motion, couples semantic segmentation and layered segmentation ({layer term}), and encourages spatial contiguity of layered segmentation using a fully-connected CRF model ({space term}).

The {\bf data term} imposes appearance constancy when corresponding pixels are visible at the same layer, and a constant penalty otherwise. It reasons about occlusions by comparing the layer assignment of corresponding pixels:
\begin{eqnarray}
\lefteqn{E_{\textrm{data}}(\mv{u}_{tk}, \! \mv{v}_{tk}, \! \mv{g}_{tk};   \mv{I}_{t}, \mv{I}_{t\!+\!1}) =} \nonumber \\
& &\sum_p  \rho_D(I_t^p \!-\!I_{t+1}^q) \delta({g}^p_{t1} \!=\!  {g}^q_{t+1, 1}) \!+\! \nonumber\\
& &\lambda_D \delta({g}^p_{t1} \!\neq\!  {g}^q_{t+1, 1}),
\end{eqnarray}
where $q\!=\!(x\!+\!{{u}^p_{tk}}, y\!+\!{v}^p_{tk})$ denotes the corresponding pixel according to the motion for pixel $p$, for every pixel in the image, $\rho_D$ is a robust penalty function, and $\lambda_D$ is a constant penalty for occluded pixels and pixels of different objects. The indicator function $\delta(x)$ is $1$ if the expression $x$ is true, and $0$ otherwise.

The {\bf motion term} encodes two assumptions. First, neighboring pixels should have similar motion if they belong to the same layer.
Second, pixels from each layer $k$ should share a global motion model $\bar{\mv{u}}(\Theta_{tk})$, where $\Theta_{tk}$ are parameters that change over time and depend on the object class $k$:
%
%
%
\begin{eqnarray}
\label{eq:motion1}
\lefteqn{E_{\textrm{motion}}(\mv{u}_{tk},\! \mv{v}_{tk}, \! \mv{g}_{tk}, \Theta_{tk}) \! =\! }  \nonumber \\ 
& & \sum_{p} \sum_{\spaq \in \mathcal{N}_p} \rho({u}^p_{tk} \!-\! {u}^\spaq_{tk})\delta({g}^p_{tk}\!=\! {g}^\spaq_{tk}) \!+\! \nonumber \\
& &  \lambda_{\textrm{aff}} \sum_p \rho_{\textrm{aff}} ({u}^p_{tk} \!-\! \bar{u}^p(\Theta_{tk})) 
\end{eqnarray}
where the set $\mathcal{N}_p$ contains the four nearest neighbors of pixel $p$. The motion term for the vertical flow field $\mv{v}_t$ is defined similarly.

The {\bf time term} encourages corresponding pixels over time to have the same layer label
\begin{align}
E_{\textrm{time}}(\mv{u}_{tk},\! \mv{v}_{tk}, \! \mv{g}_{t k}, \mv{g}_{t\!+\!1 k}) \!  =\!  \sum_p \delta( {g}_{tk}^p \!\neq\! {g}_{t\!+\!1 k}^q),
\end{align}
where $q$ is the corresponding pixel at the next frame for $p$ according to the motion $(\mv{u}_{tk}, \mv{v}_{tk})$.

The {\bf space term} encourages spatial contiguity of layer segmentation:
\begin{align}
E_{\textrm{space}}(\mv{g}_{tk}) = \sum_p \sum_{\spaq\!\neq p} w^{p}_\spaq \delta(g^p_{tk} \!\neq\! g^\spaq_{tk}),
\end{align}
where the weight $ w^{p}_\spaq$ is the same as in Sun et al.~\cite{Sun:CVPR:2013}. This term fully connects each pixel with all other pixels in the localized region.
In our implementation, we modify the approach in \cite{Sun:CVPR:2013} and apply this, not over the whole frame, but over a detected object region.

The major difference from Sun \etal~\cite{Sun:CVPR:2013} 
is that we have a semantic segmentation for the foreground and this segmentation is usually reasonably good.
Consequently we define a new {\bf coupling term}, $E_{\textrm{layer}}$, that enforces similarity between the foreground layer segmentation and the semantic segmentation:
%
\begin{equation}
E_{\textrm{layer}}(\mv{g}_{tk}; \hat{\mv{g}}_{tk}) = \sum_p \ \delta \big(  {g}^p_{tk} \!\neq\!  \hat{g}^p_{tk} \big),
\end{equation}
where $\hat{g}_t$ is the segmentation mask of the foreground Thing. 

\paragraph{Initialization and optimization.}
The layer method requires an initialization of the foreground region $\mv{g}$, an initial flow $\hat{\mv{u}}$, and parametric motions of both layers $\bar{\mv{u}}(\Theta)$.

The initial flow is typically inaccurate at the boundaries and we do not want this to corrupt the initialization.
Consequently we compute the initial affine motion ignoring the pixels close to the object boundary both in the background and foreground. 
We then optimize Eq.~\ref{eq:fg} using the method in \cite{Sun:CVPR:2013}.
This refines the flow of each layer and the segmentation (\Fig~\ref{fig:method}).
The segmentation is quite accurate because it uses backward and forward flow and image evidence with the fully connected model in the region (see  \cite{Sun:CVPR:2013}).
The method~\cite{Sun:CVPR:2013} uses heuristics to reason about depth ordering. Here we use the class category to decide the depth ordering and assume that Things are always foreground.

\subsection{Composing the Flow Field}

\begin{figure}[t]
\centerline{\includegraphics[width=\columnwidth]{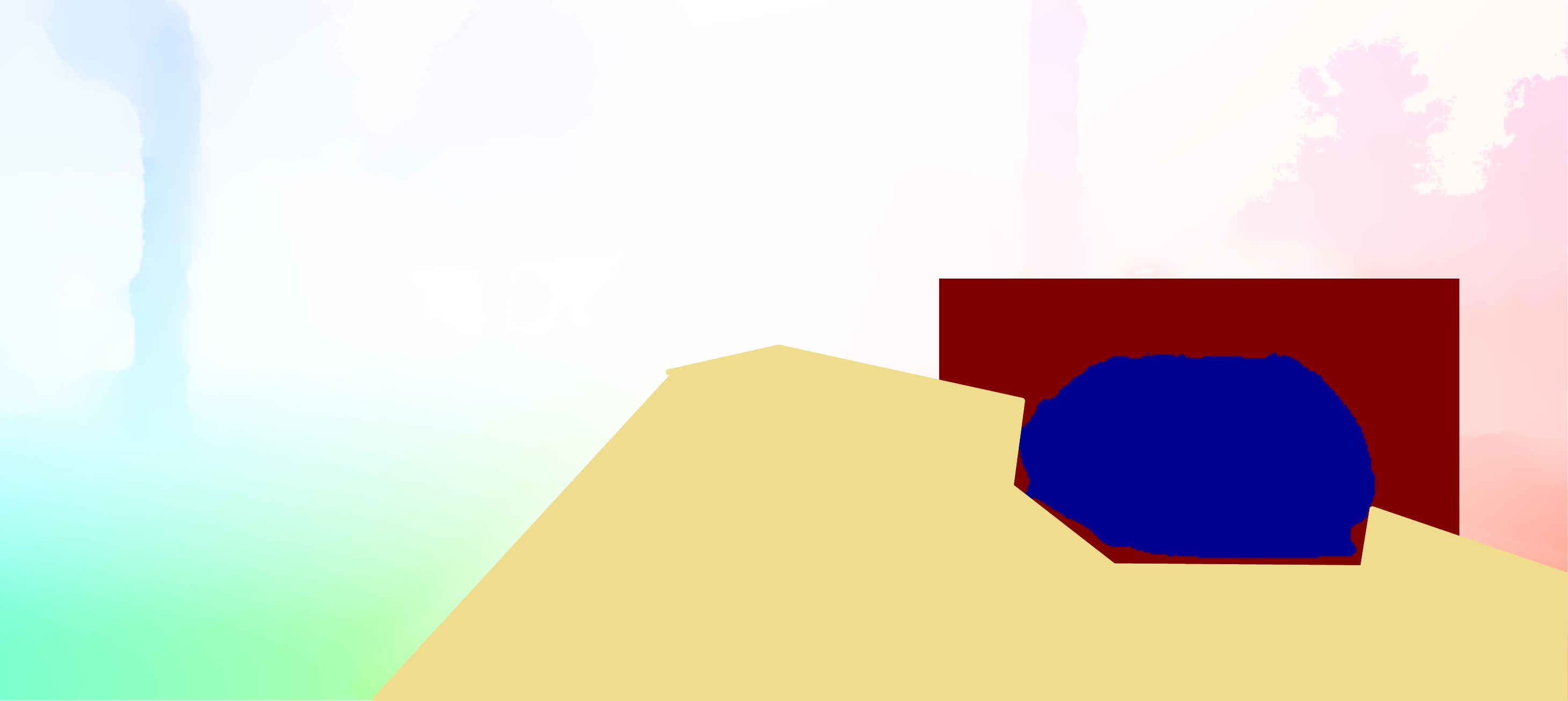}}
\caption{{\bf Compositing the flow.} The motion of Stuff, Planes (yellow) and regions around Things (red and blue) is composited to produce the final flow estimation.
}
\label{fig:composing}
\end{figure}

Each Plane and Stuff region gives exactly one flow value per pixel.
If these pixels are not occluded by a localized layer, then their flow becomes the final flow value.
The localized layers estimate the flow of the foreground and background pixels within an object region.
These regions may extend over Plane and Stuff regions, giving multiple possible flow values for these overlapped pixels.
We select a single value for each such pixel as follows (\Fig~\ref{fig:composing}). 
The foreground flow is directly pasted onto the flow field (blue region). 
When the background region of a localized layer overlaps a Plane, we keep the planar motion (yellow region).
When the background overlaps a Stuff region, we take a weighted average of the Stuff flow and the layer flow (red region).
The weight for the layer flow is high near the foreground and decays to zero at the region boundary.
Thus we favor the layered flow estimate near the foreground because it tends to be more accurate at boundaries. 
We found this approach faster and better than FusionFlow~\cite{FusionFlow}. 

\section{Experiments}
\label{sec:experiments}

\noindent
We test our Semantic Optical Flow (SOF) method in two different datasets: natural Youtube sequences and KITTI 2015~\cite{Menze2015CVPR}. Standard optical flow benchmarks do not contain the variety of objects that a semantic segmentation method can recognize. Thus, we collected a suite of natural videos from YouTube, containing objects of the Pascal VOC classes that move. Although there is no ground truth to provide a quantitative analysis, the difference of quality is clearly visible in planar regions and at motion boundaries. 
All sequences will be made publicly available \cite{website}.
In addition, we test our method on the KITTI 2015 dataset, where existing semantic segmentation methods perform reasonably well. We do not include results on the Sintel dataset because semantic segmentation does not produce reasonable results. This is probably due to the fact that the statistics of synthetically generated images are different from those of natural images, like the ones in the enriched Pascal VOC dataset. We tried training the same network using the Sintel training set (manually annotated), and we found that the network did not perform well, presumably due to a shortage of training data. In the Middlebury dataset~\cite{Baker:2011} the semantic segmentation results produce mostly the `unknown' class, or they correspond to classes without a specific motion model (\ie building), or they are very small regions and we do not consider them. Thus, on Middlebury our results are identical to the initial flow (DiscreteFlow) in all sequences but in one, where our accuracy is 0.004 better. 

\subsection{KITTI 2015}

\begin{figure*}
\centering
\includegraphics[width=\textwidth]{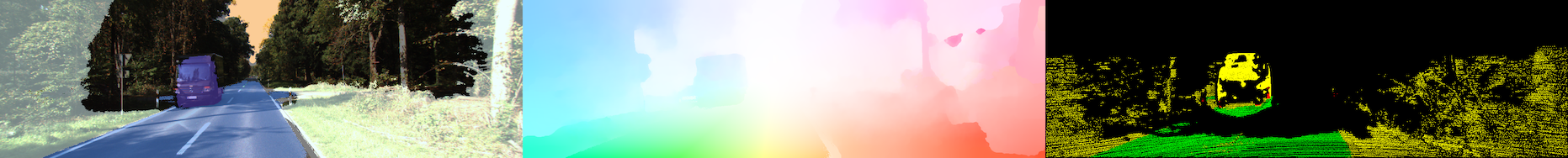}
\includegraphics[width=\textwidth]{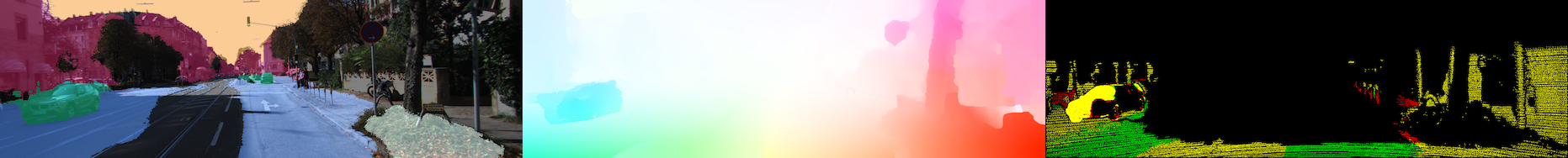}\\
\includegraphics[width=\textwidth]{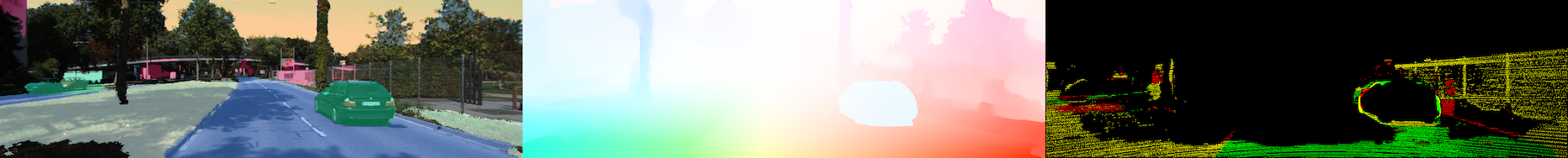}\\
\includegraphics[width=\textwidth]{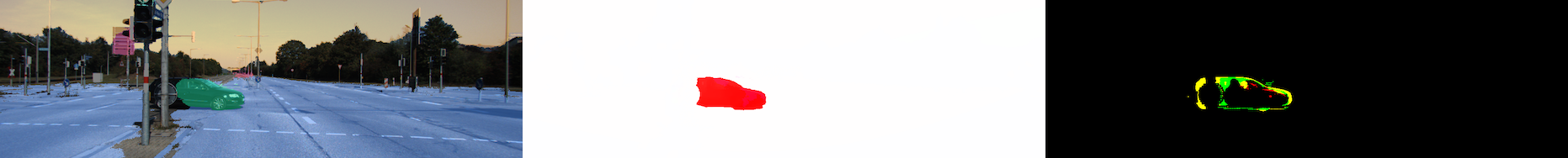}\\
\caption{{\bf Examples of Semantic Optical Flow on KITTI 2015.} From left to right: Initial segmentation; Optical flow estimation from SOF; Comparison of outliers between DiscreteFlow and SOF (black pixels indicate neither algorithm produced an outlier in that location, yellow pixels indicate both methods produced an outlier, green pixels indicate DiscreteFlow was incorrect SOF was correct, and red pixels indicate DiscreteFlow was correct but SOF was not). Notice that much of the gain from SOF is on the road, especially at occluded regions, and on the areas close to cars. }
\label{fig:kitti_full}
\end{figure*}

\begin{figure}
\centering
\includegraphics[width=0.49\textwidth]{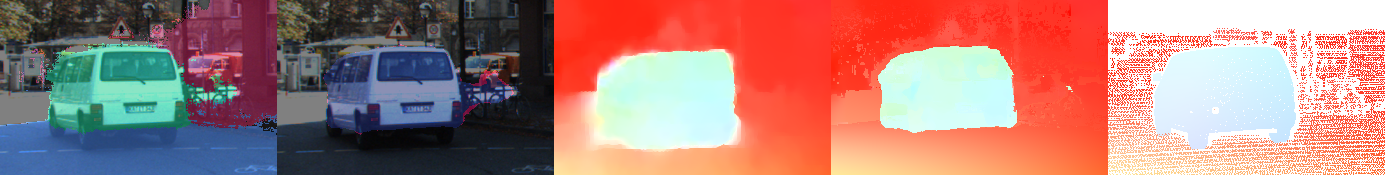}
\includegraphics[width=0.49\textwidth]{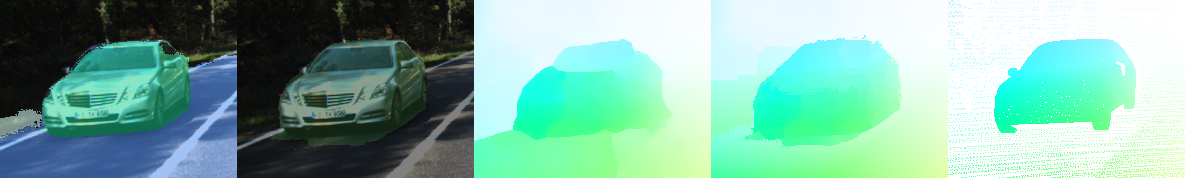}\\
\includegraphics[width=0.49\textwidth]{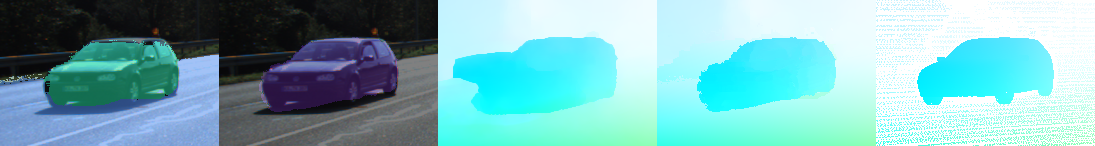}\\
\includegraphics[width=0.49\textwidth]{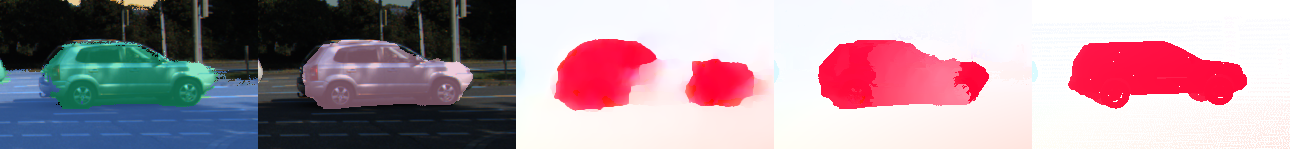}\\
\includegraphics[width=0.49\textwidth]{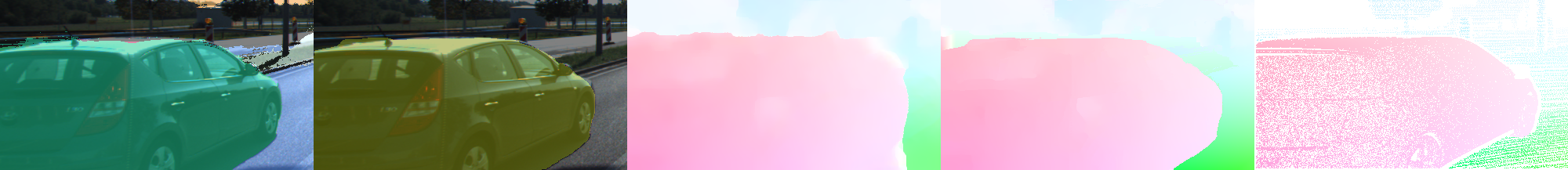}\\
\includegraphics[width=0.49\textwidth]{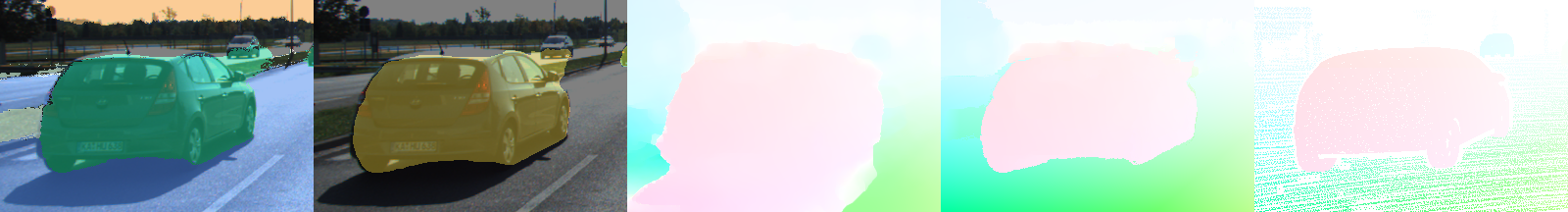}\\
\includegraphics[width=0.49\textwidth]{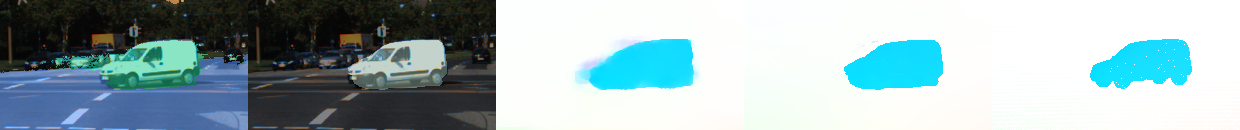}\\
\caption{{\bf Comparison of details recovered by Semantic Optical Flow.} From left to right: Initial segmentation; SOF segmentation; Optical flow estimation from DiscreteFlow; Optical flow estimation from SOF; Ground truth flow.}
\label{fig:kitti_boxes}
\end{figure}

We quantitatively evaluate our method on the KITTI 2015 benchmark (Fig.~\ref{fig:kitti_full}) using $T=2$ frames as input.
A numerical comparison between DiscreteFlow, FullFlow \cite{ChenCVPR16}, and our method is shown in Table~\ref{table:kitti_test}. Our method significantly reduces the overall percentage of outliers compared with DiscreteFlow (from 22.38\% to 16.81\%). The improvements mainly come from 1) our refined motion for the Planes; and 2) correctly interpolated motion for the occluded background regions. Figure~\ref{fig:kitti_boxes} shows several examples where our method fixes large errors of the foreground cars in the initial DiscreteFlow results.

\begin{figure}
\centering
\includegraphics[width=0.49\textwidth]{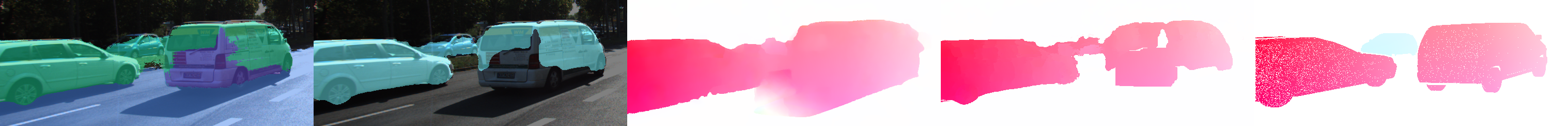}
\caption{{\bf Failure case.} From left to right: Initial segmentation; SOF segmentation; Flow estimation from DiscreteFlow; Flow estimation from SOF; Ground truth flow. Our layered method assumes two dominant motions in the region, failing if there are more than two motions. }
\label{fig:kitti_failure}
\end{figure}

Our method has a slightly higher percentage of outliers in the foreground region.
This reveals a tradeoff between segmentation and flow accuracy.
The more we restrict the foreground to affine motion, the better the segmentation but the worse the flow estimate.
Also our method only assumes two major motions are present in the detected region, and it may fail when the assumption does not hold (\Fig~\ref{fig:kitti_failure}). 
This is due to our segmentation method giving a class segmentation and grouping multiple objects together.
To address this, we either need instance-level segmentation of Things or 
a formulation that deals with more than two layers \cite{Sun:2012:LSOT}.

The execution time of our method depends on the size of the image, the number of objects, and the size of these. An upper bound for the total time is 6 minutes for a frame of KITTI 2015. Specifically, the initial semantic segmentation takes 10 seconds, the initial motion estimation from
DiscreteFlow takes 3 minutes, the motion of Planes takes 2 seconds, and the motion of Things depends on the size of the object, but takes on average 1-2 minutes. 

\begin{table}[t]
\begin{center}
{\small
    \setlength{\tabcolsep}{0.1cm}
    \begin{tabular} { r  r  r  r  r  r  r}
    Method  & \specialcell{\bf Fl-all \\ (All px)}  &  \specialcell{Fl-bg \\ (All px)} &   \specialcell{Fl-fg \\ (All px)}  &  \specialcell{\bf Fl-all \\ (Nocc)}  &  \specialcell{Fl-bg \\ (Nocc)} &   \specialcell{Fl-fg \\ (Nocc)}  \\
    \hline
    Full & 24.26\% & 23.09\% & 30.11\% & 15.35 \% & 12.97\% & 26.10\% \\
    Discrete & 22.38\%  &  21.53\% & 26.68\%  & 12.18\%  & 9.96\% & 22.17\%  \\
    SOF & 16.81\%  &  14.63\% & 27.73\% & 10.86\%  &  8.11\% & 23.28\% \\
       \end{tabular}
       \vspace{-5ex}
}
\end{center}
    \caption{Results for the test set of KITTI 2015. We compare with DiscreteFlow \cite{Menze2015GCPR} and FullFlow \cite{ChenCVPR16}, which is the next most accurate published monocular method.}
    \label{table:kitti_test}
\end{table}

\subsection{Natural Sequences.}

\begin{figure*}
\centering
\includegraphics[width=\textwidth]{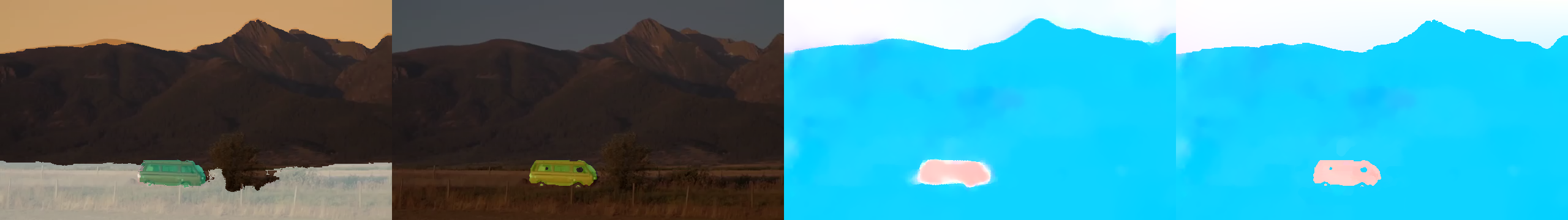}
\includegraphics[width=\textwidth]{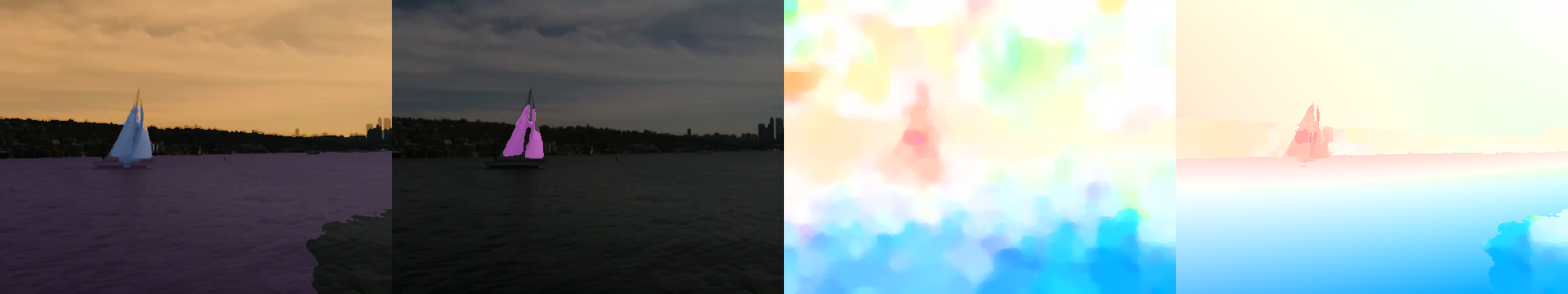}\\
\includegraphics[width=\textwidth]{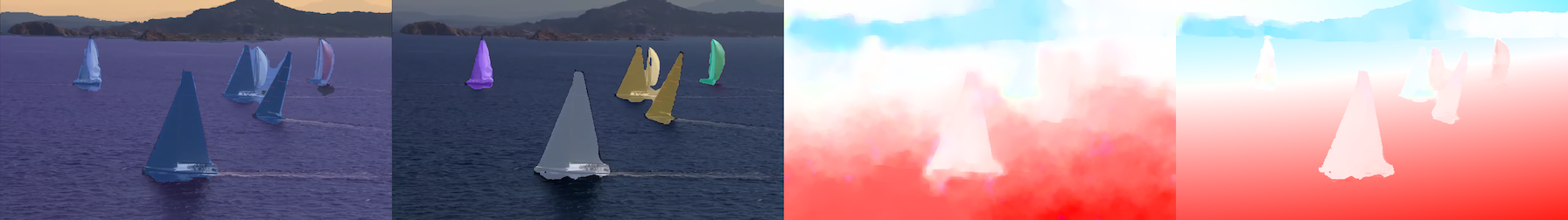}\\
\includegraphics[width=\textwidth]{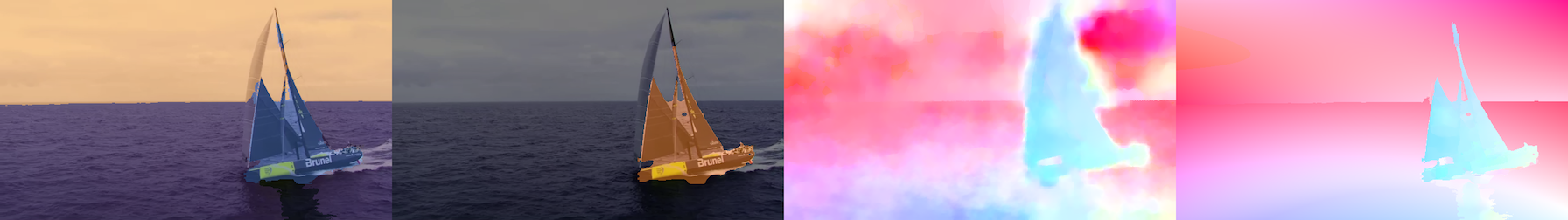}\\
\includegraphics[width=\textwidth]{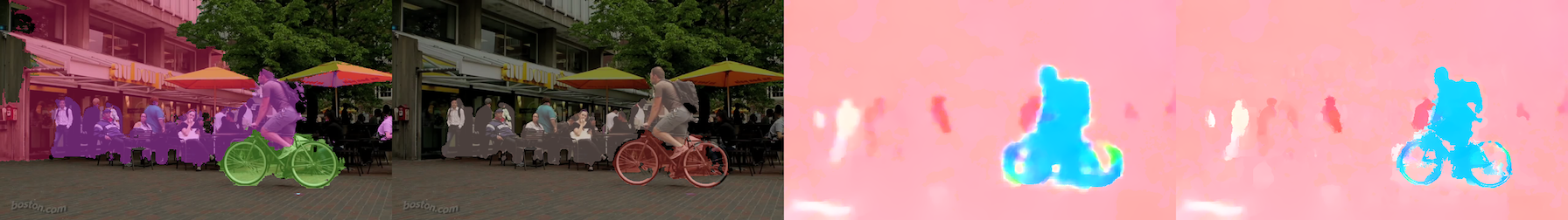}\\
\includegraphics[width=\textwidth]{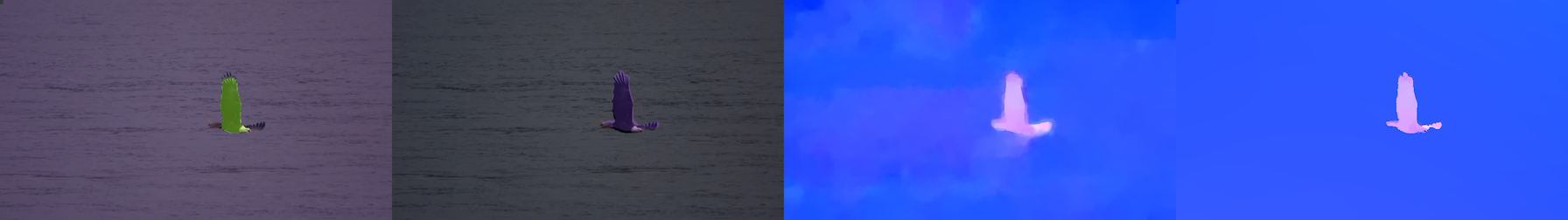}\\
\includegraphics[width=\textwidth]{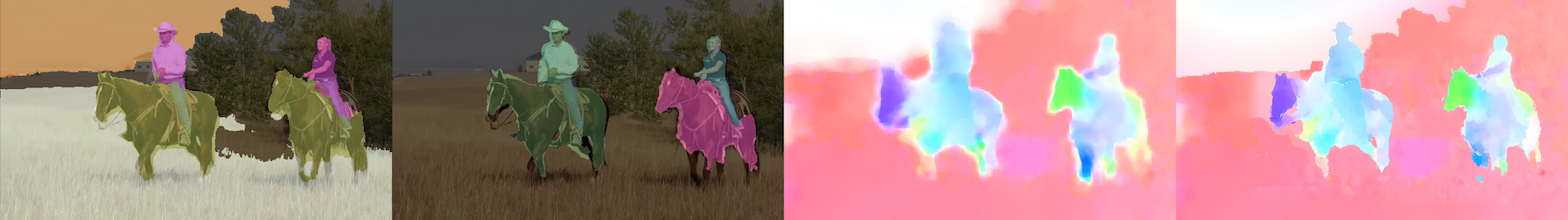}
\caption{{\bf Qualitative analysis of Semantic Optical Flow.} We show a few representative examples from the YouTube dataset. From left to right: Initial segmentation, SOF segmentation, optical flow estimation from DiscreteFlow, optical flow estimation from SOF.
More examples can be found at \cite{website}.
}
\label{fig:nat_seqs}
\end{figure*}

Figure~\ref{fig:nat_seqs} shows examples on natural sequences downloaded from YouTube.
We estimate the flow using non-overlapping 5-frame subsequences.
Our method improves over the state-of-the-art optical flow estimation method. It corrects errors in large planar regions and produces more accurate motion boundaries. It is also able to refine the semantic segmentation, especially at object boundaries and in thin regions.
These results demonstrate the benefits of our approach when reliable semantic segmentation is available.

\section{Conclusion and Future work}
\label{sec:conclusion}

We have defined a method for using semantic segmentation to improve
optical flow estimation.
Our semantic optical flow method uses object class labels to determine the appropriate motion
model to apply in each region.
We classify a scene into Things, which move independently, Planes,
which are large, roughly planar regions, and Stuff, which is everything
else.
We focus on the estimation of Things using a localized layer model in
which we only apply layered optical flow in constrained regions around
objects of interest.
We introduce a novel constraint to prefer layered segmentations that
resemble our semantic segmentation.
A key insight is that a detected object region is likely to contain
at most two motions and the object is likely to be in front.
We show that using motion we are able to visually improve the segmentation,
sometimes dramatically.
We tested the method on the KITTI-2015 flow benchmark and have the
lowest error of any monocular method by a significant margin at the
time of writing.
We also tested on a wide range of other videos containing more varied
classes and see clear qualitative improvement in terms of flow and
segmentation. This work confirms the benefit of using high quality segmentation for
optical flow and for exploiting knowledge of the class labels in
estimating flow.
This opens several doors for future work.
In particular, it may be possible to formulate our localized layer model  as a
single objective function and optimize it as such; this may improve
results further.
Additionally it would be useful, but challenging, to integrate flow
estimation with semantic segmentation.
Flow information may even help with class recognition in addition to
segmentation.

\noindent
{\bf Acknowledgments.} We thank Martin Kiefel and Jonas Wulff for
their help and insight. We thank Raquel Urtasun for helpful comments on the manuscript.

\pagebreak

{\small
\bibliographystyle{ieee}

}

\end{document}